\documentclass[twoside,11pt]{article}

\usepackage{jmlr2e}

\usepackage{color}
\usepackage{amsmath}
\usepackage{xspace}
\usepackage[dvipsnames]{xcolor}
\usepackage{multirow}
\usepackage{booktabs}
\usepackage{caption}
\usepackage{subcaption}

\usepackage[scaled]{beramono}
\usepackage{listings}
\usepackage{wrapfig}
\usepackage{capt-of}
\usepackage{caption}
\usepackage{fontawesome}
\usepackage{pifont}
\usepackage{array}
\usepackage{wasysym}
\usepackage{xr}
\usepackage{xr-hyper}

\hypersetup{citecolor=black, linkcolor=black}

\DeclareFixedFont{\ttb}{T1}{txtt}{bx}{n}{9}
\DeclareFixedFont{\ttm}{T1}{txtt}{m}{n}{9}

\definecolor{deepblue}{rgb}{0,0,0.5}
\definecolor{deepred}{rgb}{0.6,0,0}
\definecolor{deepgreen}{rgb}{0,0.5,0}

\newcommand{\params}{\boldsymbol{\theta}}

\newcommand{\skscope}{\textsf{skscope}\xspace}
\newcommand{\abess}{\textsf{abess}\xspace}

\newcommand{\sklearn}{\textsf{scikit-learn}\xspace}
\newcommand{\numpy}{\textsf{numpy}\xspace}

\newcommand{\black}{\textsf{Black}\xspace}

\newcommand{\nlopt}{\textsf{nlopt}\xspace}
\newcommand{\jax}{\textsf{jax}\xspace}
\newcommand{\cvxpy}{\textsf{cvxpy}\xspace}

\newcommand{\na}{{\ding{55}}}

\newcommand{\forwardsolver}{\textsf{ForwardSolver}\xspace}
\newcommand{\scopesolver}{\textsf{ScopeSolver}\xspace}
\newcommand{\graspsolver}{\textsf{GraspSolver}\xspace}
\newcommand{\htpsolver}{\textsf{HTPSolver}\xspace}
\newcommand{\ihtsolver}{\textsf{IHTSolver}\xspace}
\newcommand{\fobasolver}{\textsf{FoBaSolver}\xspace}
\newcommand{\ompsolver}{\textsf{OMPSolver}\xspace}
\newcommand{\pdassolver}{\textsf{PDASSolver}\xspace}
\newcommand{\gurobi}{\textsf{GUROBI}\xspace}

\usepackage{lastpage}
\jmlrheading{25}{2024}{1-\pageref{LastPage}}{11/23; Revised
8/24}{9/24}{23-1574}{Zezhi Wang, Junxian Zhu, Xueqin Wang, Jin Zhu, Huiyang Peng, Peng Chen, Anran Wang, Xiaoke Zhang}
\ShortHeadings{title}{Wang, Zhu, Wang, Zhu, Peng, Chen, Wang and Zhang}

\ShortHeadings{Fast Sparsity-Constrained Optimization for Everyone}{Wang, Zhu, Wang, Zhu, Peng, Chen, Wang and Zhang}
\firstpageno{1}

\begin{document}

\title{\skscope : Fast Sparsity-Constrained Optimization in Python}
\author{\name Zezhi Wang\textsuperscript{1}, Junxian Zhu\textsuperscript{2} \email homura@mail.ustc.edu.cn, junxian@nus.edu.sg \\
  \name Xueqin Wang\textsuperscript{1}, Jin Zhu\textsuperscript{3} \email wangxq20@ustc.edu.cn, J.Zhu69@lse.ac.uk \\
  \name Huiyang Peng\textsuperscript{1}, Peng Chen\textsuperscript{1} \email \{kisstherain, chenpeng1\}@mail.ustc.edu.cn \\
  \name Anran Wang\textsuperscript{1}, Xiaoke Zhang\textsuperscript{1} \email \{wanganran, zxk170091\}@mail.ustc.edu.cn \\
  \textsuperscript{1} \addr Department of Statistics and Finance/International Institute of Finance, School of Management, University of Science and Technology of China, Hefei, Anhui, China \\
  \textsuperscript{2} \addr Saw Swee Hock School of Public Health, National University of Singapore, Singapore \\
  \textsuperscript{3} \addr Department of Statistics, London School of Economics and Political Science, London, UK 
}
\editor{Sebastian Schelter}
\maketitle
\begingroup\renewcommand\thefootnote{*}
\footnotetext{Zezhi Wang and Junxian Zhu contributed equally. Xueqin Wang is the corresponding author.}
\endgroup

\begin{abstract}
  Applying iterative solvers on sparsity-constrained optimization (SCO) requires tedious mathematical deduction and careful programming/debugging that hinders these solvers' broad impact. In the paper, the library \skscope is introduced to overcome such an obstacle. With \skscope, users can solve the SCO by just programming the objective function. The convenience of \skscope is demonstrated through two examples in the paper, where sparse linear regression and trend filtering are addressed with just four lines of code. More importantly, \skscope's efficient implementation allows state-of-the-art solvers to quickly attain the sparse solution regardless of the high dimensionality of parameter space. Numerical experiments reveal the available solvers in \skscope can achieve up to \textsf{80x} speedup on the competing relaxation solutions obtained via the benchmarked convex solver. \skscope is published on the Python Package Index (PyPI) and Conda, and its source code is available at: \url{https://github.com/abess-team/skscope}.
\end{abstract}


\section{Introduction}
Sparsity-constrained optimization (SCO) seeks for the solution of
\begin{equation}\label{eqn:sco}
  \arg\min_{\params} f(\params), \textup{ s.t. } \| \params \|_0 \leq s,
\end{equation}
where $f: \mathbb{R}^p \to \mathbb{R}$ is a differential objective function, $\|\params\|_0$ is the number of nonzero entries in $\params$, and $s$ is a given integer. 
Such optimization covers a wide range of problems in machine learning, such as compressive sensing, trend filtering, and graphical models. 
SCO is extremely important for the ML community because it naturally reflects Occam's razor principle of simplicity. 
Nowadays, active studies prosper solvers for the SCO \citep{cai2011orthogonal, foucart2011hard, beck2013sparsity, bahmani2013greedy, liu2014forward,shen2017tight, yuan2020dual, zhou2021global, zhu2023scope}. 
In spite of the successful progress, two reasons still hinder the application of SCO in practice. The first reason may be that recruiting these solvers for general objective functions requires tedious mathematics derivations that impose highly non-trivial tasks for board users. Next, but even worse, users have to program for the complicated mathematics derivations and algorithmic procedures by themselves, which is another thorny task for general users. 
Finally and fatally, there is no publicly available software implementing these solvers for general SCO problems.

In this paper, we propose a Python library for the SCO to fill this gap such that users can conduct these solvers with minimal mathematics and programming skills. This library, called \skscope, implements the prototypical procedures of well-known iterative solvers for general objective functions. More importantly, \skscope leverages the powerful automatic differentiation (AD) to conduct the algorithmic procedures without deriving and programming the exact form of gradient or hessian matrix \citep{rall1981automatic, baydin2018automatic}. There is no doubt that AD is the cornerstone of the computational framework of deep learning \citep{paszke2017autodiff}; and now, it is first used for efficiently solving SCO problems. 

The \skscope can run on most Linux distributions, macOS, and Windows 32 or 64-bit with Python (version $\geq 3.9$), and can be easily installed from PyPI and Conda\footnote{PyPI: \url{https://pypi.org/project/skscope}, and Conda: \url{https://anaconda.org/conda-forge/skscope}}.
We offer a website\footnote{\url{https://skscope.readthedocs.io}} to present \skscope's features and syntax. To demonstrate the versatility of \skscope, it has been applied to more than 25 machine learning problems\footnote{\url{https://skscope.readthedocs.io/en/latest/gallery}}, covering linear models (for example, quantile regression and robust regression), survival analysis (for example, Cox proportional hazard model, competitive risk model), graphical models, trend filtering and so on. It relies on GitHub Actions for continuous integration. The \black style guide keeps the source Python code clean without hand-formatting.
Code quality is assessed by standard code coverage metrics.
The coverages for the Python packages at the time of writing were over 95\%. The dependencies of \skscope are minimal and just include the standard Python library such as \numpy, \sklearn; additionally, two powerful and well-maintained libraries, \jax and \nlopt \citep{frostig2018compiling, johnson2014nlopt}, are used for obtaining AD and solving unconstrained nonlinear optimization, respectively. The source code is distributed under the MIT license.

\section{Overview of Software Features}
\begin{wraptable}{r}{0.4\textwidth}
  \vspace*{-18pt}
  \linespread{1.05}\selectfont
  \setlength{\tabcolsep}{3.3pt}
  \centering
  {\small
  \begin{tabular}{c|c}
    \toprule
    Solver & Reference \\
    \midrule
    \forwardsolver & \citet{marcano2010feature} \\
    \ompsolver & \citet{cai2011orthogonal} \\
    \htpsolver & \citet{foucart2011hard} \\
    \ihtsolver & \citet{beck2013sparsity} \\
    \graspsolver & \citet{bahmani2013greedy} \\
    \pdassolver  & \citet{wen2020bess} \\
    \fobasolver  & \citet{liu2014forward} \\
    \scopesolver & \citet{zhu2023scope} \\
    \bottomrule
  \end{tabular}}
  \vspace*{-10pt}
  \caption{Supported SCO solvers.}
  \label{tab:support-solver}
\end{wraptable}
\skscope provides a comprehensive set of state-of-the-art solvers for SCO listed in Table~\ref{tab:support-solver}. For each implemented solver, once it receives the objective function programmed by users, it will leverage AD and an unconstrained nonlinear solver to get the ingredients to perform iterations until a certain convergence criterion is met. The implementation of each solver has been rigorously tested and validated through reproducible experiments, ensuring its correctness and reliability. Detailed reproducible results can be found on the public GitHub repository\footnote{\url{https://github.com/abess-team/skscope-reproducibility}}.

Besides, \skscope introduces two generic features to broaden the application range. Specifically, \skscope enables the SCO on group-structured parameters and enables pre-determining a part of non-sparse parameters. Moreover, \skscope allows using information criteria or cross-validation for selecting the sparsity level, catering to the urgent needs of the data science community. Warm-start initialization is supported to speed up the selection. In terms of computation, \skscope can transparently run on the GPU/TPU and is compatible with the just-in-time compilation provided by the \jax library. This enables efficient computing of AD, resulting in the acceleration of all solvers. Typically, \skscope maintains computational scalability as state-of-the-art regression solvers like \abess \citep{zhu2022abess} while it possesses capability in solving general SCO problems (see Table~\ref{tab:abess-skscope}). Furthermore, \skscope enables optimized objective functions implemented with sparse matrices\footnote{Sparse matrices are recommended primarily for memory-intensive scenarios.} to save memory usage. Finally, as a factory for machine learning methods, the \skscope continuously supplies \sklearn compatible methods (listed in Table~\ref{tab:ml-interfaces} in Appendix), which allows practitioners to directly call them to solve practical problems.

\section{Usage Examples}
An example of compressing sensing with \graspsolver is demonstrated in Figure~\ref{fig:usage-cs}. From the results in lines 16–17, we witness that \graspsolver correctly identifies the effective variables and gives an accurate estimation. More impressively, the solution is easily obtained by programming 4 lines of code. 
\begin{figure}[htbp]
  \begin{tiny}
    \vspace*{-3pt}
    \begin{lstlisting}[language=Python]
import numpy as np
import jax.numpy as jnp
from skscope import GraspSolver   ## the gradient support pursuit solver
from sklearn.datasets import make_regression
## generate data
x, y, coef = make_regression(n_features=10, n_informative=3, coef=True)
print("Effective variables: ", np.nonzero(coef)[0], 
      "coefficients: ", np.around(coef[np.nonzero(coef)[0]], 2))
def ols_loss(params):      ## define loss function
    return jnp.linalg.norm(y - x @ params)
## initialize the solver: ten parameters in total, three of which are non-zero
solver = GraspSolver(10, 3)  
params = solver.solve(ols_loss) 
print("Estimated variables: ", solver.get_support(), 
      "estimated coefficients:", np.around(params[solver.get_support()], 2))
>>> Effective variables:  [3 4 7] coefficients:  [ 9.71 19.16 13.53]
>>> Estimated variables:  [3 4 7] estimated coefficients: [ 9.71 19.16 13.53]
\end{lstlisting}
    \vspace*{-0.5cm}
  \end{tiny}
  \caption{Using the \skscope for compressive sensing.}\label{fig:usage-cs}
  \vspace*{-3pt}
\end{figure}

\begin{figure}[htbp]
  \vspace*{-0.24cm}
  \begin{subfigure}[b]{0.7\textwidth}
    \begin{tiny}
\begin{lstlisting}[language=Python]
import numpy as np
import jax.numpy as jnp
import matplotlib.pyplot as plt
from skscope import ScopeSolver
np.random.seed(2023)
# observed data, random walk with normal increment:
x = np.cumsum(np.random.randn(500)) 
def tf_objective(params):
	return jnp.linalg.norm(data - jnp.cumsum(params))
solver = ScopeSolver(len(x), 10)
params = solver.solve(tf_objective)
plt.plot(x, label='observation', linewidth=0.8)
plt.plot(jnp.cumsum(params), label='filtering trend')
plt.legend(); plt.show()
\end{lstlisting}
    \end{tiny}
  \end{subfigure}
  \hspace*{-20pt}
  \begin{subfigure}[b]{0.29\textwidth}
    \includegraphics[width=1.2\linewidth]{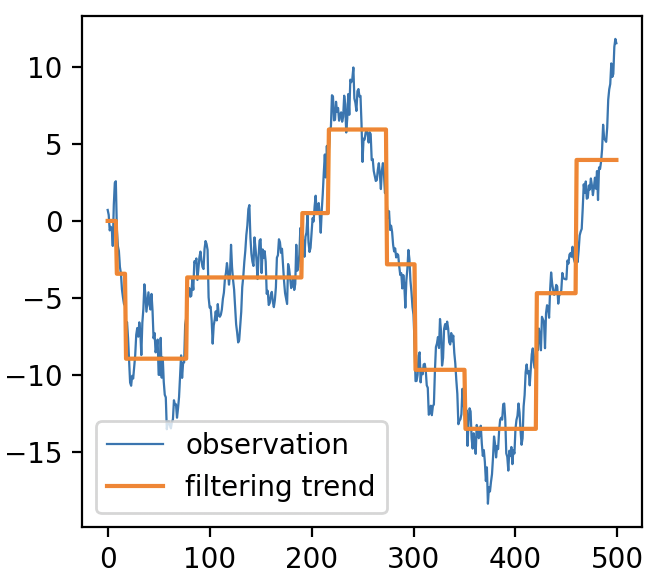}
  \end{subfigure}
  \vspace*{-0.26cm}
  \caption{Using the \skscope for trend filtering.}\label{fig:usage-tf}
\end{figure}

Figure~\ref{fig:usage-tf} presents for filtering trend via \scopesolver, serving as a non-trivial example because the dimensionality of parameters is hundreds. From~\ref{fig:usage-tf} shows that the solution of \scopesolver captures the main trend of the observed data. In this case, 6 lines of code help us attain the solution. Even more impressively, for a concrete SCO problem\footnote{\url{https://skscope.readthedocs.io/en/latest/gallery/GeneralizedLinearModels/poisson-identity-link.html}} with parameters of order $O(10^4)$ and an objective function involving matrices of size $O(10^8)$, solvers in \skscope can tackle the problem in less than two minutes on a personal laptop.

\section{Performance}\label{sec:performance}

We conducted a comprehensive comparison among the sparse-learning solvers employed in \skscope and two alternative approaches. The first competing approach solves~\eqref{eqn:sco} by recruiting the widely-used mixed-integer optimization solver, \gurobi\footnote{\textsf{TimeLimit} is set to 1000. Note that optimization may not immediately stop upon hitting \textsf{TimeLimit}.}. We compare this approach assuming the optimal $s$ of~\eqref{eqn:sco} is known and present the results in Table~\ref{table:acc_time}. The second approach utilizes the $\ell_1$ relaxation of~\eqref{eqn:sco}, implemented using the open-source solver, \cvxpy \citep{diamond2016cvxpy}. The comparison with \cvxpy assumes the optimal $s$ of \eqref{eqn:sco} is unknown and searches with information criteria. The corresponding results are reported in Table~\ref{tab:select-sparsity}. These comparisons covered a wide range of concrete SCO problems and were performed on a Ubuntu platform with Intel(R) Xeon(R) Silver 4210 CPU @ 2.20GHz and 64 RAM. Python version is 3.10.9. 

Table~\ref{table:acc_time} shows that \skscope not only achieves highly competitive support-set selection accuracy but also has a significantly lower runtime than \gurobi. Table~\ref{tab:select-sparsity} reveals that, in terms of support-set selection, all solvers in \skscope generally have a higher precision score while maintaining a competitive recall score, leading to a higher F1-score. The results indicate that \skscope outperforms \cvxpy in overall selection quality. Furthermore, as shown in Tables~\ref{table:acc_time} and \ref{tab:select-sparsity}, \skscope has a desirable support set selection quality when $f(x)$ is non-convex or non-linear where \cvxpy or \gurobi may fail. In terms of computation, \skscope generally demonstrated significant computational advantages against \cvxpy and \gurobi, exhibiting approximately \textsf{1-80x} speedups on \cvxpy and \textsf{30-1000x} speedups on \gurobi. Among the solvers in \skscope,  \scopesolver and \fobasolver have particularly promising results in support set selection, with \scopesolver achieving speedups of around \textsf{1.5x-7x} compared to \fobasolver.  

\section{Conclusion and Discussion}
\skscope is a fast Python library for solving general SCO problems. It offers well-designed and user-friendly interfaces such that users can tackle SCO with minimal knowledge of mathematics and programming. Therefore, \skscope must have a broad application in diverse domains. Future versions of \skscope plan to support more iterative solvers for the SCO  \citep[e.g.,][]{zhou2021global} so as to establish a benchmark toolbox/platform for the SCO.

\acks{We would like to thank the editor and the two referees for their constructive comments and valuable suggestions, which have substantially improved both the library and the paper. We also thank Kangkang Jiang, Junhao Huang, and Yu Zheng for their insightful discussions.

Wang's research is partially supported by National Natural Science Foundation of China grants No. 72171216, 12231017, 71921001, and 71991474. 
}

\hypersetup{hidelinks}

\appendix

\setcounter{table}{0}
\renewcommand{\thetable}{A\arabic{table}}

\section{Additional Tables.}

\begin{table}[htbp]
\centering
\begin{tabular}{c|c|cc|cc}
\toprule
\multirow{4}{*}{Linear}     & \multirow{2}{*}{Method}  & \multicolumn{2}{c|}{$n=500, p=1000, s=10$}     & \multicolumn{2}{c}{$n=5000, p=10000, s=100$} \\
\cline{3-6} 
& & Accuracy           & Runtime            & Accuracy          & Runtime          \\ 
\cline{2-6} 
& \abess               & 1.00 (0.00)    & 0.01 (0.00)      & 1.00 (0.00)   &  2.19 (0.22) \\
& \scopesolver         & 1.00 (0.00)    &  0.23 (0.03)     & 1.00 (0.00)   &  14.09 (1.28)    \\
\midrule
\multirow{4}{*}{Logistic}     & \multirow{2}{*}{Method}  & \multicolumn{2}{c|}{$n=500, p=1000, s=10$}     & \multicolumn{2}{c}{$n=5000, p=10000, s=100$} \\
\cline{3-6} 
& & Accuracy           & Runtime            & Accuracy          & Runtime          \\ 
\cline{2-6} 
& \abess               & 0.99 (0.03)    & 0.02 (0.00)      & 1.00 (0.00)   &  6.81 (3.32) \\
& \scopesolver         & 0.99 (0.03)    & 0.26 (0.03)     & 1.00 (0.00)   &  24.12 (5.83)    \\
\midrule
\multirow{4}{*}{NNLS}     & \multirow{2}{*}{Method}  & \multicolumn{2}{c|}{$n=500, p=1000, s=10$}     & \multicolumn{2}{c}{$n=5000, p=10000, s=100$} \\
\cline{3-6} 
& & Accuracy           & Runtime            & Accuracy          & Runtime          \\ 
\cline{2-6} 
& \abess               & 1.00 (0.00)    & 0.01 (0.00)      & 1.00 (0.00)   &  2.26 (0.10) \\
& \scopesolver         & 1.00 (0.00)    &  0.22 (0.02)     & 1.00 (0.00)   &  14.80 (1.00)    \\
\bottomrule
\end{tabular}
\caption{Comparison between \abess and \skscope on linear regression,  logistic regression models, and non-negative least square (NNLS) estimation. The datasets are generated using the \textsf{make\_glm\_data} function implemented in the \abess package. For all tasks, the non-zero coefficients are randomly chosen from $\{1, \ldots, p\}$. The mean metrics are computed over 10 replications with standard deviation in parentheses. For large problems, \skscope runs \textsf{3x}-\textsf{7x} slower than \abess. However, \skscope can handle the same problem scale as \abess. For instance, if \abess can process a dataset of size $n \times p$, \skscope can handle a dataset of size $(n/3) \times (p/3)$. For smaller problems, although \skscope is much slower than \abess, it solves problems in less than 0.3 seconds, providing immediate results for users.}
\label{tab:abess-skscope}
\end{table}

\begin{table}
  \begin{center}
    \begin{tabular}{c|c}
      \toprule
      \textsf{skmodel} & Description \\
      \midrule
      \textsf{PortfolioSelection} & Construct sparse Markowitz portfolio\\
      \textsf{NonlinearSelection} & Select relevant features with nonlinear effect \\
      \textsf{RobustRegression} & A robust regression dealing with outliers\\
      \textsf{MultivariateFailure} & Multivariate failure time model in survival analysis \\
      \textsf{IsotonicRegression} & Fit the data with an non-decreasing curve\\
      \bottomrule
    \end{tabular}
  \end{center}
  \vspace*{-10pt}
  \caption{Some application-oriented interfaces implemented in the module \textsf{skmodel} in \skscope.}
  \label{tab:ml-interfaces}
\end{table}  

\begin{table}
  \vspace*{-0.35cm}
  \begin{center}
    {\footnotesize
    \begin{tabular}{c|cc|cc|cc}
   	\toprule
\multirow{2}{*}{Method} & \multicolumn{2}{c|}{Linear regression}     & \multicolumn{2}{c|}{Logistic regression} & \multicolumn{2}{c}{Robust feature selection} \\ \cline{2-7} 
& Accuracy           & Runtime            & Accuracy          & Runtime           & Accuracy           & Runtime             \\ \hline
\ompsolver               & 1.00(0.01)    & 2.45(0.68)      & 0.91(0.05)   &  1.66(0.67)     & 0.56(0.17)   &  0.73(0.14)      \\
\ihtsolver               & 0.79(0.04)    & 3.42(0.88)      & 0.97(0.03)   &  1.06(0.67)     & 0.67(0.07)   &  0.89(0.22)      \\
\htpsolver               & 1.00(0.00)    & 4.14(1.25)      & 0.84(0.05)   &  2.37(0.92)     & 0.91(0.05)   &  5.00(0.94)      \\
\graspsolver             & 1.00(0.00)    & 1.16(0.38)      & 0.90(0.08)   &   12.70(8.20)   & 1.00(0.00) &  0.50(0.25)      \\
\fobasolver              & 1.00(0.00)    &  11.70(2.90)    & 0.92(0.06)   &  6.31(2.15)     & 0.98(0.08)   &  3.37(0.66)      \\
\scopesolver             & 1.00(0.00)    &  2.11(0.70)     & 0.94(0.04)   &  3.24(2.67)     & 0.98(0.09)   &  1.86(0.55)      \\
\gurobi                  & 1.00(0.00)    & 1009.94(0.66)   & \na          &  \na            & \na          &  \na           \\
\midrule
\multirow{2}{*}{Method} & \multicolumn{2}{c|}{Trend filtering} & \multicolumn{2}{c|}{Ising model}    & \multicolumn{2}{c}{Nonlinear feature selection}   \\ \cline{2-7} 
& Accuracy           & Runtime            & Accuracy          & Runtime           & Accuracy           & Runtime             \\ \hline
\ompsolver         &    0.86(0.03)   &   1.77(0.57)      & 0.98(0.03)   &  2.86(0.86)    & 0.77(0.09)    & 11.53(3.61)     \\
\ihtsolver         &     0.08(0.00)  &   0.76(0.28)      & 0.96(0.05)   &  3.24(1.43)    & 0.78(0.09)    &  6.37(2.32)     \\
\htpsolver         &    0.47(0.03)   &   0.71(0.23)      & 0.97(0.03)   &  5.26(2.03)    & 0.78(0.09)    & 10.82(7.86)     \\
\graspsolver       &    0.78(0.12)   &   0.95(0.38)      & 0.99(0.01)   &  1.02(0.44)    & 0.78(0.08)    &  7.34(2.75)     \\
\fobasolver        &      1.00(0.00) &   8.27(1.66)      & 1.00(0.01)   & 11.59(3.55)    & 0.77(0.09)    &  31.26(8.80)     \\
\scopesolver       &    0.98(0.02)   &   4.73(1.13)      & 1.00(0.01)   &  1.69(0.67)    & 0.77(0.09)    &    8.60(2.70)     \\
\gurobi            &   1.00(0.00)    & 1013.16(0.62)     & \na         &  \na          &  0.79(0.08)     & 1003.88(1.53)     \\
\bottomrule
   \end{tabular}
    }
  \end{center}
  \vspace*{-10pt}
  \caption{\label{table:acc_time}The numerical experiment results on six specific SCO problems. Accuracy is equal to $|\operatorname{supp}\{\params^*\} \cap \operatorname{supp}\{\params\}| / |\operatorname{supp}\{\params^*\}|$ and the runtime is measured by seconds. The results are the average of 100 replications, and the parentheses record standard deviation. Robust (or nonlinear) variable selection is based on the work of \citep{xueqin2013robust} (or \citep{hsiclasso2014yamada}). \gurobi: version 10.0.2; \cvxpy: version 1.3.1; \skscope: version 0.1.8. \na: not available.} 
\end{table}

\begin{table}[!t]
\begin{center}
\setlength{\tabcolsep}{2.8pt}
\renewcommand{\arraystretch}{1.25}
{\scriptsize
\begin{tabular}{c|cccc|cccc}
\toprule
\multirow{2}{*}{Method} & \multicolumn{4}{c|}{Linear regression}     & \multicolumn{4}{c}{Logistic regression} \\ \cline{2-9} 
& Recall      & Precision   & F1-score  & Runtime            & Recall      & Precision    & F1-score  & Runtime             \\ \hline
\ompsolver   &  0.90 (0.28)  & 1.00 (0.01) & 0.92 (0.24) & 69.29 (3.74)  & 0.91 (0.06) & 0.91 (0.07) & 0.91 (0.06) & 64.78 (3.63)    \\
\ihtsolver   &  0.90 (0.28)  & 1.00 (0.01) & 0.92 (0.24) & 14.27 (1.59)  & 0.92 (0.06) & 0.93 (0.07) & 0.93 (0.06) & 39.38 (1.87)    \\
\htpsolver   &  0.93 (0.25)  & 1.00 (0.01) & 0.94 (0.21) & 60.76 (20.04) & 0.91 (0.06) & 0.91 (0.07) & 0.91 (0.06) & 68.33 (31.86)   \\
\graspsolver &  0.66 (0.45)  & 1.00 (0.00) & 0.69 (0.42) & 15.10 (1.94)  & 0.98 (0.04) & 0.94 (0.07) & 0.96 (0.05) & 86.07 (9.21)    \\
\fobasolver  &  0.92 (0.26)  & 1.00 (0.00) & 0.93 (0.23) & 234.09 (9.39) & 0.91 (0.06) & 0.93 (0.07) & 0.92 (0.07) & 206.86 (13.51)  \\
\scopesolver &  0.92 (0.26)  & 1.00 (0.00) & 0.93 (0.23) & 9.38 (0.46)   & 0.93 (0.05) & 0.96 (0.06) & 0.95 (0.05) & 7.29 (1.03)     \\
\cvxpy       &  1.00 (0.00)  & 0.43 (0.03) & 0.60 (0.03) & 55.44 (10.9)  & 1.00 (0.01) & 0.42 (0.03) & 0.59 (0.03) & 867.76 (203.71) \\
\midrule
\multirow{2}{*}{Method} & \multicolumn{4}{c|}{Robust feature selection} & \multicolumn{4}{c}{Trend filtering} \\ \cline{2-9} 
& Recall      & Precision   & F1-score   & Runtime    & Recall          & Precision    & F1-score    & Runtime        \\ \hline
\ompsolver   & 0.98 (0.09) & 0.66 (0.12) & 0.66 (0.12) &   69.01 (4.49)  & 0.53 (0.04) & 0.79 (0.05) & 0.63 (0.04) & 143.94 (8.72)  \\
\ihtsolver   & 1.00 (0.02) & 0.78 (0.09) & 0.78 (0.09) &   65.99 (4.43)  & 0.53 (0.04) & 0.79 (0.05) & 0.63 (0.04) & 6.59 (0.28)    \\
\htpsolver   & 1.00 (0.00) & 0.80 (0.11) & 0.80 (0.11) & 667.02 (36.75)  & 0.53 (0.04) & 0.80 (0.05) & 0.64 (0.04) & 33.53 (1.56)   \\
\graspsolver & 1.00 (0.00) & 1.00 (0.00) & 1.00 (0.00) &   23.91 (3.54)  & 0.63 (0.17) & 0.71 (0.21) & 0.66 (0.18) & 21.46 (13.19)  \\
\fobasolver  & 0.99 (0.06) & 0.98 (0.09) & 0.98 (0.09) & 246.68 (19.46)  & 0.84 (0.09) & 1.00 (0.00) & 0.91 (0.06) & 322.80 (17.12) \\
\scopesolver & 1.00 (0.00) & 0.99 (0.06) & 0.99 (0.06) &    9.28 (0.77)  & 0.66 (0.09) & 0.90 (0.05) & 0.76 (0.07) & 17.81 (1.25)   \\
\cvxpy       &        \na  &        \na  &    \na      &           \na   & 0.76 (0.11) & 0.31 (0.05) & 0.44 (0.07) & 67.69 (10.28)  \\
\midrule
\multirow{2}{*}{Method} & \multicolumn{4}{c|}{Ising model} & \multicolumn{4}{c}{Nonlinear feature selection} \\ \cline{2-9} 
& Recall      & Precision  & F1-score   & Runtime         & Recall      & Precision   & F1-score    & Runtime         \\ \hline
\ompsolver   & 0.99 (0.02) & 0.92 (0.06) & 0.95(0.04) &  132.79 (9.75) & 0.79 (0.09) & 0.78 (0.12) & 0.78 (0.08) &  227.62 (59.59) \\
\ihtsolver   & 0.99 (0.02) & 0.78 (0.19) & 0.86(0.13) &   97.03 (5.35) & 0.42 (0.08) & 0.77 (0.16) & 0.54 (0.09) &  158.25 (35.93) \\
\htpsolver   & 0.99 (0.02) & 0.80 (0.19) & 0.87(0.13) &  98.18 (15.63) & 0.42 (0.08) & 0.77 (0.16) & 0.54 (0.09) &  290.92 (65.97) \\
\graspsolver & 1.00 (0.01) & 0.93 (0.06) & 0.96(0.04) &  32.80 (15.14) & 0.79 (0.09) & 0.79 (0.12) & 0.78 (0.08) &   51.72 (11.44) \\
\fobasolver  & 1.00 (0.01) & 0.93 (0.04) & 0.96(0.02) & 432.28 (23.38) & 0.79 (0.09) & 0.78 (0.12) & 0.78 (0.08) & 874.08 (195.91) \\
\scopesolver & 1.00 (0.02) & 0.93 (0.05) & 0.96(0.03) &   36.14 (1.93) & 0.79 (0.09) & 0.78 (0.12) & 0.78 (0.09) &  198.37 (43.24) \\
\cvxpy       &        \na &        \na & \na          &       \na      & 0.76 (0.09) & 0.83 (0.11) & 0.79 (0.08) & 525.64 (122.28) \\
\bottomrule
\end{tabular}}
\caption{The numerical experiment when the optimal sparsity $s$ in~\eqref{eqn:sco} is unknown and information criteria are used for selecting the optimal one. Specifically, for linear regression, special information criterion \citep{zhu2020polynomial} is used; as for logistic regression, Ising model, and non-linear feature selection, generalized information criterion \citep{zhu2023best} is employed; and for trend filtering and robust feature selection, Bayesian information criterion \citep{wen2023tf} is used. Recall: the recall score that is computed by $|\operatorname{supp}\{\params^*\} \cap \operatorname{supp}\{\params\}| / |\operatorname{supp}\{\params^*\}|$; Precision: the precision score computed by $|(\operatorname{supp}\{\params^*\})^c \cap \operatorname{supp}\{\params\}| / |(\operatorname{supp}\{\params^*\})^c|$; F1-score is computed as the harmonic mean of precision and recall.}\label{tab:select-sparsity}
\end{center}
\end{table}

\clearpage
\vskip 0.2in
\bibliography{reference}
\end{document}